\documentclass[pdflatex,sn-mathphys-num]{sn-jnl}

\usepackage{graphicx}%
\usepackage{multirow}%
\usepackage{amsmath,amssymb,amsfonts}%
\usepackage{amsthm}%
\usepackage{mathrsfs}%
\usepackage[title]{appendix}%
\usepackage{xcolor}%
\usepackage{textcomp}%
\usepackage{manyfoot}%
\usepackage{booktabs}%
\usepackage{adjustbox}%
\usepackage{algorithm}%
\usepackage{algorithmicx}%
\usepackage{algpseudocode}%
\usepackage{listings}%
\usepackage{cleveref}
\usepackage{subcaption}
\usepackage{todonotes}
\usepackage{comment}
\usepackage[dvipsnames]{xcolor}
\usepackage{tabularx}

\theoremstyle{thmstyleone}%
\theoremstyle{thmstyletwo}%

\theoremstyle{thmstylethree}%

\begin{document}

\title[REACT++: Efficient Cross-Attention for Real-Time Scene Graph Generation]{REACT++: Efficient Cross-Attention for Real-Time Scene Graph Generation}


\author*[1]{\fnm{Maëlic} \sur{Neau}}\email{maelic.neau@umu.se}

\author[1]{\fnm{Zoe} \sur{Falomir}}\email{zoe.falomir@umu.se}

\affil*[1]{\orgdiv{Department of Computing Science}, \orgname{Umeå University}, \orgaddress{\street{Universitetstorget 4}, \city{Umeå}, \postcode{907 36}, \country{Sweden}}}


\abstract{Scene Graph Generation (SGG) is a task that encodes visual relationships between objects in images as graph structures. SGG shows significant promise as a foundational component for downstream tasks, such as reasoning for embodied agents. To enable real-time applications, SGG must address the trade-off between performance and inference speed. However, current methods tend to focus on one of the following: (1) improving relation prediction accuracy, (2) enhancing object detection accuracy, or (3) reducing latency, without aiming to balance all three objectives simultaneously.
To address this limitation, we build on the powerful Real-time Efficiency and Accuracy Compromise for Tradeoffs in Scene Graph Generation (REACT) architecture and propose REACT++, a new state-of-the-art model for real-time SGG. By leveraging efficient feature extraction and subject-to-object cross-attention within the prototype space, REACT++ balances latency and representational power.
REACT++ achieves the highest inference speed among existing SGG models, improving relation prediction accuracy without sacrificing object detection performance. Compared to the previous REACT version, REACT++ is 20\% faster with a gain of 10\% in relation prediction accuracy on average. The code is available at \url{https://github.com/Maelic/SGG-Benchmark}.}




\maketitle

\section{Introduction}
\label{sec1}
Scene Graph Generation (SGG) is the task of generating a structured graph representation from visual inputs. Given an image, visual regions corresponding to objects of interest are extracted, and relations between these objects are predicted in the form of \texttt{<subject, predicate, object>} triplets.
The set of these triplets forms a directed acyclic graph which can be interpreted as a structured representation of the scene.
This task has garnered increasing attention recently thanks to its utility in various downstream applications, such as Visual Question Answering (VQA) \cite{hudsonGQANewDataset2019,damodaran2021understanding,liangGraghVQALanguageGuidedGraph2021}, image captioning \cite{yang2022reformer,wangRoleSceneGraphs2019,nguyenDefenseSceneGraphs2021a}, reasoning of an embodied agent \cite{gadre2022continuous,li2022embodied} but also in robotics \cite{neau2023defense,amodeo2022og,wang2025spatial}.
However, we observe a significant imbalance between the progress in the SGG task itself and its adoption in downstream tasks \cite{changComprehensiveSurveyScene2023a}. 
One of the reasons is the lack of low-cost approaches that could power applications with real-time constraints, such as visual understanding of a robotic agent \cite{neau2023defense,amodeo2022og}. Because SGG is a task that combines Object Detection (OD) and Relation Prediction (RelPred), a good trade-off between accuracy in these two parameters is also required.

In this work, we explore the bottlenecks in SGG to find an optimal combination of OD accuracy, RelPred accuracy, and latency. SGG methods can be Two-Stage (TS) \cite{zellers_neural_2018, tang_learning_2019, zheng2023prototype} or One-Stage (OS) \cite{li_sgtr_2022, cong_reltr_2022,im2024egtr} approaches. The former uses a combination of an object detector and a relation predictor to model objects and relations in a sequence. The latter uses a single-stage pipeline to infer relations and object proposals directly from the image features. As a result, one-stage approaches are often more efficient for real-time processing \cite{im2024egtr}. 
However, they are insufficiently accurate in the OD task \cite{li_sgtr_2022,im2024egtr}. 
In contrast, two-stage approaches train the object detector and relation predictor sequentially, leading to improved performance in object detection since the object detector is frozen during relation training. Two-stage approaches also possess the advantage of being modular, as the same object detection checkpoint can be used as a backbone for multiple SGG models.

In the previous version of this work \cite{Neau_2025_BMVC}, the REACT model has been proposed, combining a YOLO backbone for real-time OD and feature extraction, and efficient prototype learning for RelPred in a Decoupled Two-Stage (DTS) architecture with efficient inference thanks to Dynamic Candidate Selection (DCS). 
While the original REACT approach was able to provide substantial improvements in latency (2.7x time faster than previous work) and in OD accuracy (+ 58\%), the relation prediction component still suffered from the inherent biases of TS approaches. By analyzing the latency of the relation head of REACT we observed three critical bottlenecks: (1) the feature extraction is inefficient and consumes an important part of the parameters ; (2) the relation head is not leveraging global scene information and (3) the interactions between prototypes in the latent space are symmetric, hurting the ability of the model to learn efficient representation of the directionality of relations.

In this work, we extend the DTS architecture of REACT with a new feature extractor and relation head, aiming at achieving competitive performance with previous work while ensuring low latency. We first propose a direct feature extraction approach to replace the slow ROI Align algorithm with a more efficient approach, based on YOLO's one-stage architecture. We describe this new algorithm as the Detection-Anchored Multi-Scale Pooling (DAMP).
Next, we propose to use a compact visual representation of the scene, modeled through a dedicated Attention-based Intra-scale Feature Interaction (AIFI) component \cite{zhao2024detrs}, for efficient global context pooling.
Finally, we propose a new prototype learning approach for relation modeling, with dedicated subject and object cross-attention layers to encode the asymmetry between their representations in the latent space. In this approach, we also infuse spatial information as Rotary Position Embedding (RoPE) \cite{su2024roformer} in the cross-attention layer to help the model interpret spatial biases. As a result, we name this new relation head component CARPE for Cross-Attention Rotary Prototype Embedding.

In the following, we provide a summary of our contributions:
\begin{itemize}
    \item \textbf{DAMP}: We propose DAMP (Detection-Anchored Multi-Scale Pooling), a new simple pooling algorithm for one-stage object detectors such as YOLO. This new algorithm beats traditional ROI Align in both latency and accuracy in the SGG task.
    \item \textbf{Global Context}: Inspired by RT-DETR \cite{zhao2024detrs}, we propose to use a low-cost AIFI module to extract global information of the scene, which complements the subject/object representations.
    \item \textbf{CARPE}: We proposed CARPE (Cross-Attention Rotary Prototype Embedding), a new relation head for Scene Graph Generation based on cross attention between subject/object and predicate prototype representations. Thanks to the encoding of spatial information as position embedding, we alleviate the need for an additional spatial feature extractor \cite{Neau_2025_BMVC}.
\end{itemize}

The rest of the paper is organized as follows.
Section \ref{sec:rw} describes related work and the research gap this work fills.
Section \ref{sec:bottlenecks} explores the bottlenecks in SGG by discussing architectural issues in the feature extraction and relation prediction components of previous work.
Section \ref{sec:RT-SGG} describes the new REACT++ architecture. We dive into the main components: Decoupled Two-Stage (DTS), the DAMP feature extractor, the global context AIFI, the CARPE relation head, and the Dynamic Candidate Selection (DCS) algorithm for inference.
Section \ref{sec:Experiments-results} explains our experimentation and analyzes our results on three different datasets.
Section \ref{sec:ablation} presents the results of our ablation studies for the DAMP, AIFI, and the DCS.
And finally, our conclusions and future work are detailed in Section \ref{sec:conclusion}.

\section{Related Work}
\label{sec:rw}

This section is divided in two parts: first, we give an overview of the task of Scene Graph Generation and related state of the art. Second, we contextualize our work by presenting related approaches in Real-Time Scene Graph Generation.

\subsection{Scene Graph Generation}

The modern definition of the task of SGG takes its roots in 2017 with the introduction of the first large-scale Scene Graph dataset: Visual Genome (VG) \cite{krishna_visual_2017}. Thanks to extensive crowd-sourcing effort, VG has collected extensive annotations for more than 100K images, kick starting efforts in deep learning for relationship modeling. At the same time, Lu et. al. \cite{lu2016visual} introduced the use of Language Priors in the form of word embeddings fused with visual features for visual relationship prediction. Subsequent approaches such as Neural Motifs (2018) \cite{zellers_neural_2018} and Graph R-CNN \cite{yang_graph_2018} will introduced the use of Faster-RCNN \cite{ren2015faster} as an object detection backbone for the task, defining a \textbf{Two-Stage} pipeline for the task. These methods and subsequent ones will keep language priors as additional signal for relation modeling. Neural Motifs and Graph-RCNN also focus on context learning and inter-dependencies of relations to boost relation modeling. VCTree (2019) \cite{tang_learning_2019} and GPS-Net (2020) \cite{lin_gps-net_2020} both introduced deeper context reasoning and global-to-local information sharing for enhanced performance. In 2020, Tang et. al. \cite{tang_unbiased_2020} raised concerns about biases in the data distribution of the VG dataset, making the task a long-tail problem. They pointed language biases that can make SGG models collapse to coarse predicates like "on" instead of more fine-grained such as "sitting on" and proposed a solution based on causal intervention. Later work will build on this idea by proposing other de-biasing strategies such as Predicate-Correlation Perception Learning \cite{yan_pcpl_2020} or Group Collaborative Learning \cite{dong_stacked_2022}. The Visual Genome dataset also received criticism for sparse and noisy annotations \cite{li_devil_2022,yang_panoptic_2022}, which led to new datasets to be proposed such as the Panoptic Scene Graph (PSG) dataset \cite{yang_panoptic_2022} or the IndoorVG dataset \cite{neau2023defense}. At the same time we have seen the emergence of Transformers-based architectures for direct triplet modeling to alleviate the slow and computationally intensive dense matching of traditional approaches \cite{cong_reltr_2022, li_sgtr_2022}. These approaches will be referred as \textbf{One-Stage} approaches to the task. In recent years, we have seen a shift in the SGG community with approaches focusing on de-biasing and long-tail learning by sticking to the Two-Stage design \cite{zheng2023prototype,liu2025relation} and approaches targeting efficiency with recent advances in Transformers architecture using the One-Stage approach \cite{im2024egtr,hao2025bctr}.

\subsection{Real-Time SGG}

{\bf Two-stage SGG approaches} typically use Faster-RCNN (\figureautorefname{} \ref{fig:stage1_penet}) with a VGG-16 \cite{xu_scene_2017, yang_graph_2018} or ResNeXt-101 backbone for object detection and feature extraction, as seen in Neural-Motifs \cite{zellers_neural_2018}, VCTree \cite{tang_learning_2019}, Transformer \cite{tang_unbiased_2020}, GPS-Net \cite{lin_gps-net_2020}, and PE-NET \cite{zheng2023prototype} (\figureautorefname{} \ref{fig:stage2_penet}). In these approaches, object classes probabilities given by Faster-RCNN are refined during relationship learning, via a dedicated classifier \cite{zellers_neural_2018, lin_gps-net_2020} or by aggregating features from neighboring regions \cite{yang_graph_2018}. However, this design can hinder the performance of SGG models on the task of object detection. For our REACT++ architecture, we choose to effectively \textit{decouple} the Object Detector and Relationship Predictor by removing the dedicated classifier and keeping original class probabilities.
In addition, Faster-RCNN has a large backbone which limits its real-time use. In order to improve this, recent work \cite{jinFastContextualScene2023a} proposed a real-time SGG method based on contextual cues and relying on YOLOv5 \cite{Jocher_YOLOv5_by_Ultralytics_2020} bounding boxes to learn correlations between box coordinates and predicates—excluding visual features. However, this method relies on spatial features and neglects fine-grained relations such as the difference between $eating$ or $drinking$. Another approach \cite{jinIndependentRelationshipDetection2023} uses a Relation-aware YOLO (RYOLO) to predict relation coordinates via new anchors, matched to nearby objects. Thus, lacking subject-object context leads to poor RelPred performance (3.1\% mR@100) \cite{zellers_neural_2018}. In contrast, our REACT++ architecture proposes the use of features extracted by a YOLO model to generate representations for the subject and object nodes, making the first stage of the architecture more efficient without compromising performance on the second stage. However, the REACT approach does not tackle efficiency in the second stage, relation modeling.

{\bf One-stage SGG approaches} jointly learn object and relation representations. Sparse-RCNN \cite{teng2022structured} uses triplet queries to produce objects and relations, decoded via cascade-RCNN after CNN feature extraction. RelTr \cite{congRelTRRelationTransformer2022} and SGTR \cite{liSGTREndtoendScene2024}, built on DETR \cite{carion2020end}, generate sparse relation sets, unlike dense two-stage predictions. 
More recently,  EGTR \cite{im2024egtr} builds upon Deformable-DETR for efficient object detection, paired with graph-based relation matching. RelTr and EGTR report lower latency than two-stage models like Motifs \cite{zellers_neural_2018} and VCTree \cite{tang_learning_2019}, though these comparisons often rely on Faster-RCNN. As highlighted in recent work \cite{liSGTREndtoendScene2024}, learning both the object and relation representations at the same time can make the performance of one-stage models in OD sub-optimal in comparison to two-stage approaches.
We argue that replacing Faster-RCNN with YOLO in two-stage pipelines can achieve both lower latency than two-stage approaches and higher OD accuracy than one-stage approaches, as this paper shows next.



\section{Investigating Bottlenecks in SGG}
\label{sec:bottlenecks}

This section investigates bottlenecks in SGG regarding: 
(i) Object Detection accuracy;
(ii) Linguistic Features Priors and Benchmarks Biases;
(iii) Global vs Local Context; and
(iv) Loss Function.

\begin{figure*}[t]
    \centering
    \begin{subfigure}{0.85\textwidth}
        \centering
        \includegraphics[width=\textwidth]{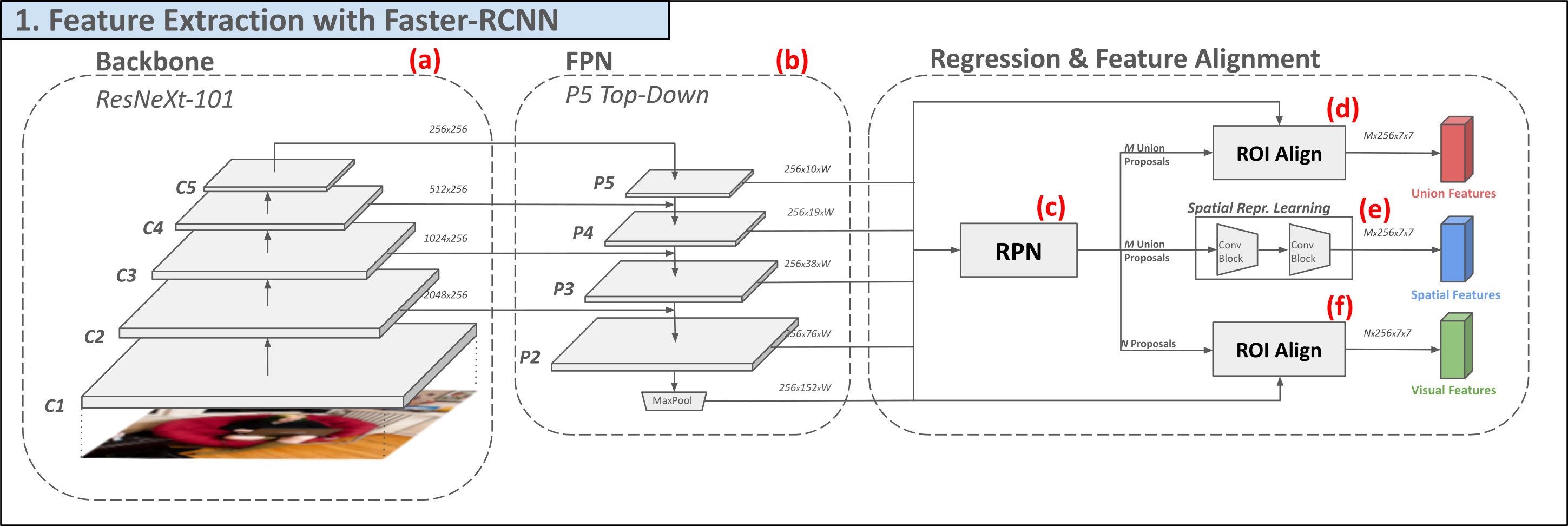}    
        \caption{PE-NET Stage 1 - Two-Stage Faster-RCNN}
        \label{fig:stage1_penet}
    \end{subfigure}
        \begin{subfigure}{0.85\textwidth}
        \centering
        \includegraphics[width=\textwidth]{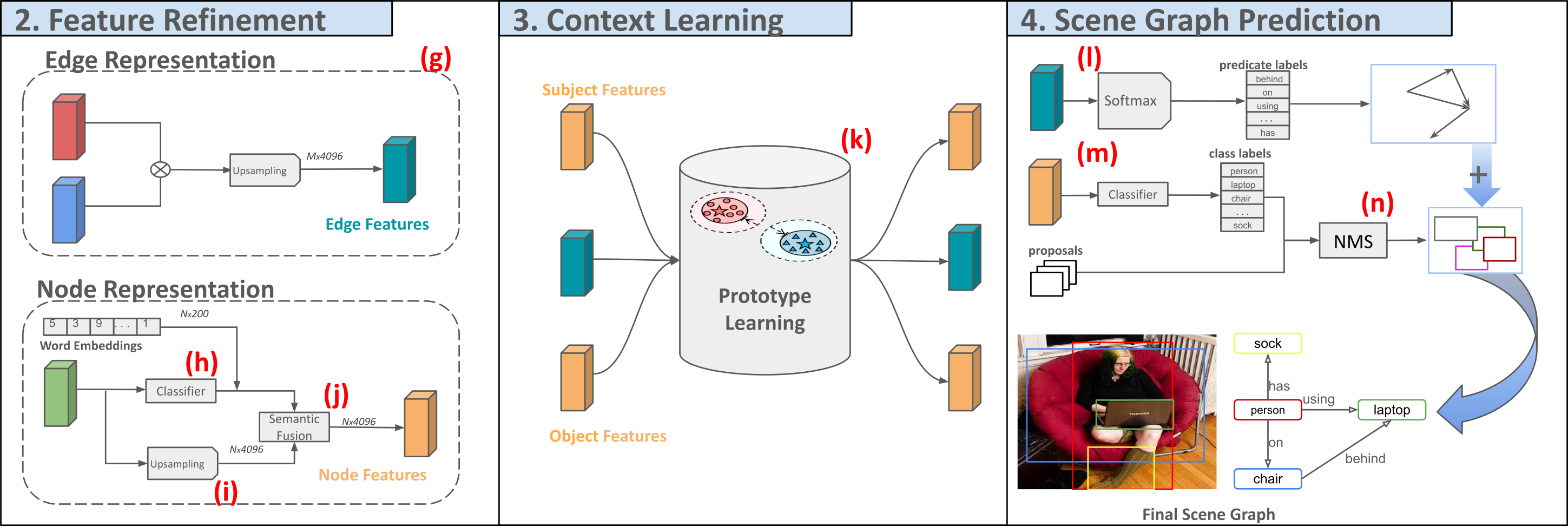}
        \caption{PENET Stage 2 - Prototype Learning (PE-NET)}
        \label{fig:stage2_penet}
    \end{subfigure}
    \caption[Pipeline of a typical SGG model]{Comparing PE-NET architecture \cite{zheng2023prototype} (top) with our REACT++ architeture (botton) in Stage 1. The dashed boxes in orange [\textcolor{orange}{\textbf{- - -}}] represent modified or added components, $\bigotimes$ denotes element-wise concatenation. }
    \label{fig:sgg_pipeline:penet}
\end{figure*}

\Cref{fig:sgg_pipeline:penet} is a representation of the PE-NET model \cite{zheng2023prototype}: a state-of-the-art two-stage SGG approach. The feature extraction (stage 1 \Cref{fig:stage1_penet}) as well as the graph prediction (module 4, \Cref{fig:stage2_penet}) are common to all two-stage architectures \cite{tang_learning_2019,tang_unbiased_2020,dong_stacked_2022,li_bipartite_2021,zellers_neural_2018,lin_gps-net_2020}.
In Stage 1, visual features are extracted by a ResNeXt-101 backbone (\Cref{fig:stage1_penet} \textbf{(a)}), and a Feature Pyramid Network (FPN) \cite{lin2017feature} is used to refine the features at different scales (\Cref{fig:stage1_penet} \textbf{(b)}). Then, a Region Proposal Network (RPN) \cite{ren2015faster} generates proposals (\Cref{fig:stage1_penet} \textbf{(c)}) which are used to align three different types of features with corresponding image regions: union features, spatial features, and visual features (\Cref{fig:stage1_penet} \textbf{(d)}, \textbf{(e)}, and \textbf{(f)}) using ROI Align \cite{ren2015faster}. 
Union and spatial features are aggregated for every possible pair of proposals and will serve as the initial edge representation to the relation prediction stage (Stage 2, \Cref{fig:stage2_penet}). A classifier extracts the class labels of each proposal (\Cref{fig:stage2_penet} \textbf{(h)}) and retrieves corresponding linguistic features from pre-trained text embeddings (GloVe \cite{pennington2014glove}). Then, the visual features are combined with the textual features to form the node representation through semantic fusion (\Cref{fig:stage2_penet} \textbf{(i)}, \textbf{(j)}). Then, prototype learning takes place (\Cref{fig:stage2_penet} \textbf{(k)}). Finally, object and predicate labels are decoded to form the final graph representation (\Cref{fig:stage2_penet} \textbf{(l)}, \textbf{(m)}, and \textbf{(n)}). Below, we discuss the main bottlenecks of this architecture for Latency and Object Detection accuracy.

\textbf{Object Detection Accuracy.} The strategy employed in the two-stage SGG involves to decode object labels twice during the relation prediction stage: once for extracting class labels to initialize text features (see \Cref{fig:stage2_penet} \textbf{(h)}), and another time to compute the final predictions (see \Cref{fig:stage2_penet} \textbf{(m)}). 
The predictions obtained from this last step depend on the context learning strategy of each model \cite{tang_learning_2019,zellers_neural_2018,lin_gps-net_2020,tang_unbiased_2020}. This architecture design is (i) redundant, as similar operations are performed twice, and (ii) might introduce significant differences in object detection accuracy in contrast to the original Faster-RCNN implementation. 

\Cref{tab:fastercnn_results} draws a comparison of the performance of the Faster-RCNN model equipped with a ResNeXt-101 feature extractor on the PSG dataset with different relation prediction heads. Note that the performance in OD drops significantly, from 1 point wrt Motifs-TDE \cite{zellers_neural_2018,tang_unbiased_2020} to almost 13 points in mAP@50 wrt GPS-NET \cite{lin_gps-net_2020} after the relation-prediction stage. These results demonstrate a clear dependency between the two stages of the pipeline, which negatively impacts model performance on the object detection (OD) task. 
A second consideration is the low performance of Faster-RCNN. Recent state-of-the-art detectors such as YOLOV8+ \cite{Jocher_Ultralytics_YOLO_2023} may be more appropriate for the task. In contrast to Faster-RCNN, the YOLO architecture is not designed for Feature Extraction. This is the main reason why the Faster-RCNN architecture is still widely used for SGG. In this work, we aim to alleviate this issue by using the YOLO architecture for both object detection AND box-specific feature extraction.

\begin{table}[t]
    \centering
    \caption{Object Detection accuracy for two-stage models, PSG dataset \cite{yang_panoptic_2022}.}
        \begin{tabular}{c|c|c}
            \hline
            Backbone & Relation Head & mAP@50 \\
            \hline
            \multirow{6}{*}{Faster-RCNN} & - & 36.4  \\
             & Neural-Motifs \cite{zellers_neural_2018} & 35.9  \\
             & PE-NET \cite{zheng2023prototype} & 35.4\\ 
             & Transformer \cite{tang_unbiased_2020} & 34.6 \\
             & VC-Tree \cite{tang_learning_2019} & 32.6  \\ 
             & GPS-NET \cite{lin_gps-net_2020} & 31.5  \\
             \hline
        \end{tabular}
    \label{tab:fastercnn_results}
\end{table}

\textbf{Global vs Local Context.}
In two-stage SGG, visual features of subject and object are aggregated locally (see \Cref{fig:stage2_penet} \textbf{(f)}), by multi-scale pooling using the ROI Pooling algorithm \cite{ren2015faster}. In addition, the features of the union boxes are also extracted (see \Cref{fig:stage2_penet} \textbf{(d)}). However, the global context of the scene is rarely used to decode relations. Recent approaches in Object Detection have shown the importance of global context to extract high-level semantic features about the scene prior to refinement and prediction \cite{zhao2024detrs}. Inspired by such approaches, we propose to use global-to-local features fusion as initial representations of relations. For relation prediction, global features can help to infer the \textit{overall context} of the scene (i.e. kitchen, mountain, beach etc...), which can help to model context-dependent predicates (such as eating, drinking, playing with, swimming etc...).

\textbf{Asymmetry of relations.} Visual relations are by definition asymmetric, as \texttt{<person, eating, pizza>} is not similar to \texttt{<pizza, eating, person>}. However, in previous work \cite{zheng2023prototype,Neau_2025_BMVC}, the same representation for each object was used to model all relations of that object, independently of its position in the relation. This is unfortunately sub-optimal, and a good modeling will enforce the asymmetry of relations in the shared embedding space, for instance, through a dedicated cross-attention layer.

\textbf{Spatial encoding.} In the Two-Stage design, spatial features are explicitly encoded using the coordinates of the union region between subject and object pairs (see \Cref{fig:stage2_penet} \textbf{(e)}). This encoding is very important, especially to learn good representation of spatial predicates such as ``on top of", ``behind", or ``below".
Since the spatial encoding is using multiple Conv blocks and an important feature size (i.e. 256 channels), it adds to the overall compute load of the model. Moreover, the spatial features are aggregated and projected into the latent space symmetrically for both the subject and object prototypes. This process can be regarded as inefficient since some prototypes should be able to look at spatial features differently depending on their position as subject or object in the relation. For instance, a person in the top-left (likely subject: ``person stands-on ...") vs. bottom-right (likely object: ``something is-above person") could have the same prototype but different geometric context. 


\textbf{Proposal Candidate Selection.} In the Two-Stage architecture, Non-Maximum Suppression (NMS) for object proposals (\Cref{fig:stage2_penet} \textbf{(n)}) is performed in stage 2.
To alleviate the lack of NMS in stage 1, a fixed set of the top proposals is traditionally considered valid node candidates for stage 2. The idea here is that by sampling a large number of proposals per image, it is easier to find valid pairs and relations. 
However, this increases the computational complexity of the graph learning. Also, the number of top proposals to select (traditionally 80 or 100) is set arbitrarily. 
Recent work \cite{zellers_neural_2018,tang_unbiased_2020,li_bipartite_2021} computes the final ranking of relations with the following score formula: \(    \theta_{rel} = \theta_{obj} * \theta_{pred} * \theta_{subj},\)
where $\theta_{pred}$ is the confidence score of a predicate, given $<subject, object>$ pair as candidate; and, $\theta_{obj}, \theta_{subj}$ are the respective confidence score of the object detector. 
This formulation assigns greater weight to the confidence scores of the subject and object compared to the predicate. As a result, many low-confidence proposals can be filtered out prior to the relation prediction stage with minimal impact on the model's overall accuracy. In theory, this process should be able to remove a consistent computational burden to the model, since relations are sampled as an $N \times (N-1)$ matrix in stage 2.

Next section introduces the REACT++ architecture, and new foundational components to solve the aforementioned problems.


\section{Toward Real-Time SGG}
\label{sec:RT-SGG}

\begin{figure*}[t]
    \centering
    \begin{subfigure}{0.85\textwidth}
        \centering
        \includegraphics[width=\textwidth]{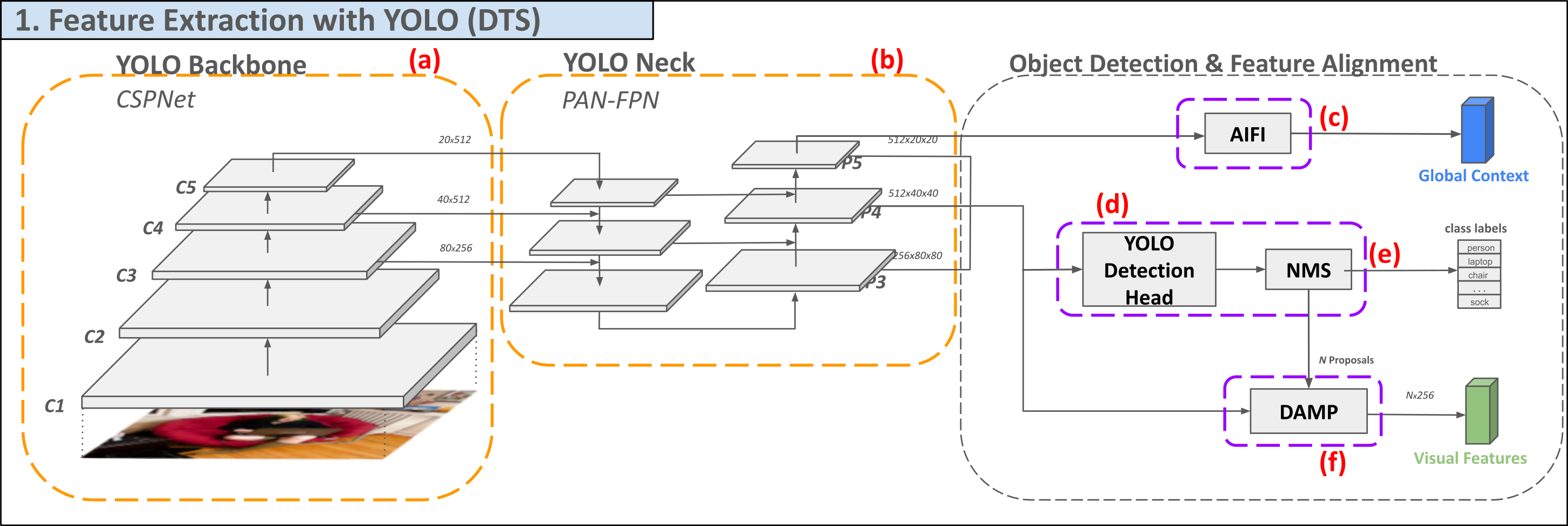}    
        \caption{Our REACT++ Stage 1 - DTS YOLO with refined feature extraction. Global context features are extracted using an AIFI block and independent visual features through our new DAMP block.}
        \label{fig:stage1_react}
    \end{subfigure}
    \begin{subfigure}{0.85\textwidth}
        \centering
        \includegraphics[width=\textwidth]{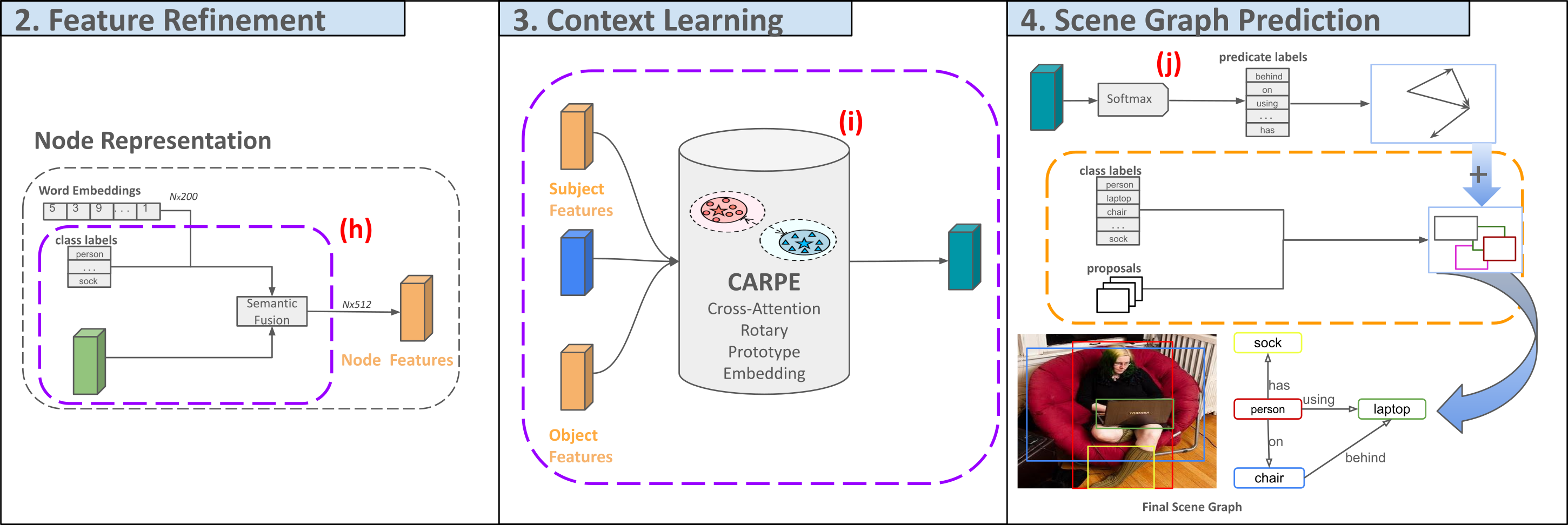}    
        \caption{Our REACT++ Stage 2 - Node representation drops the projection layer, and the context learning is reworked with the use of new cross-attention layers and RoPE encoding for low-cost directional spatial features infusion.}
        \label{fig:stage2_react}
    \end{subfigure}
    \caption[Pipeline of a typical SGG model]{Our REACT++ architecture is based on REACT \cite{Neau_2025_BMVC}. The dashed boxes in purple [\textcolor{violet}{\textbf{- - -}}] represent modified or added components to REACT. The dashed boxes in orange [\textcolor{orange}{\textbf{- - -}}] represent modified or added components to the original PE-NET (see \Cref{fig:sgg_pipeline:penet}).  $\bigotimes$ denotes element-wise concatenation.
    }
    \label{fig:sgg_pipeline:stage-2}
\end{figure*}

This work introduces REACT++, a novel architecture for real-time Scene Graph Generation (SGG). REACT++ extends the original REACT framework by adopting a decoupled two-stage design in which the Faster R-CNN detector is replaced with a YOLO-based backbone.
In addition, REACT++ incorporates a redesigned relation head that eliminates the need for multi-scale feature alignment and RoIAlign-based pooling of subject and object bounding boxes. This design choice removes a major computational bottleneck present in prior two-stage SGG methods, enabling significantly more efficient inference while preserving the advantages of two-stage modeling. Finally, REACT++ improves the expressiveness of relation modeling in prototype space through the addition of global context and dedicated cross-attention layers. 

Our architecture is depicted in \Cref{fig:sgg_pipeline:stage-2}. Following REACT, we use a DTS architecture with YOLO backbone (\Cref{fig:stage1_react}) for stage 1. Our main contributions in stage 1 are the addition of an AIFI block for global context refinement (see \Cref{fig:stage1_react} \textbf{(c)}) and our new DAMP block for visual features extraction (see \Cref{fig:stage1_react} \textbf{(f)}). In stage 2, we modified the node refinement by removing the upsampling process (see \Cref{fig:stage2_react} \textbf{(h)}). Our main contribution comes during context learning, as we make use of a new Cross-Attention with Rotary Embedding for more expressive prototype learning (see \Cref{fig:stage2_react} \textbf{(i)}). In the following, we detail each of these components.


\subsection{Decoupled Two-Stage SGG} 

To maintain the object detector accuracy during relation prediction training, our REACT++ architecture freezes the regression and the classification head of the object detector and performs NMS before the relation prediction stage. With this strategy, the object classifier in the relation prediction stage is also removed, and the class prediction from the object detector is used for the final prediction.
Thus, the relation prediction stage outputs only the pairs and associated predicates, in contrast to predicting also the object class labels. 

With the REACT++ architecture, any object detector can be used in place of Faster-RCNN, without impacting stage 2. For instance, one could consider replacing Faster-RCNN with a real-time object detector such as YOLO \cite{jinFastContextualScene2023a}. 
To take advantage of YOLO versions, we leveraged the introduced Decoupled Two-Stage architecture and replaced the ResNeXt-101 backbone with YOLO's CSPNet-based backbones \cite{Jocher_Ultralytics_YOLO_2023,wang2024yolov9} (see \Cref{fig:stage1_react} \textbf{(a)}). 
The P5-FPN of Faster-RCNN was also replaced by the YOLO PAN-FPN Neck (see \Cref{fig:stage1_react} \textbf{(b)}). Finally, we removed the RPN and detection head of Faster-RCNN and replaced them with the YOLO Detection Head (\Cref{fig:stage1_react} \textbf{(d)}). Since we now have two completely independent components, we renamed this architecture the Decoupled Two-Stage (DTS) SGG.

\subsection{DAMP: Detection-Anchored Multi-scale Pooling}

Unlike Faster R-CNN, YOLO-based architectures are not inherently designed for region-wise feature extraction, as they do not incorporate a \emph{RoI Align} \cite{ren2015faster} module. To overcome this limitation, we propose a new approach called DAPM, built upon YOLO’s grid-based representation. Below, we outline the key differences between the conventional RoI Align used in prior two-stage Scene Graph Generation (SGG) methods and our proposed DAPM.

In ROI ALign, for each of the $N$ detected objects, a $7{\times}7$ grid of bilinearly interpolated samples is extracted from the FPN feature maps at each pyramid level. These sampled features are then pooled, flattened, and aggregated to produce object-level representations. The computational cost is dominated by the bilinear interpolation operations, with complexity $O(N \times 7^2)$. During benchmarking of REACT and other two-stage models, we observed that RoI Align accounts for a substantial fraction of the total inference time, contributing up to 40\% of the latency of the relation head.

In YOLOv8 and subsequent versions, the traditional anchor-based grid formulation has been replaced with an anchor-free design. In this architecture, every spatial feature from the three feature maps of the PAN-FPN Neck is directly processed by the detection head. Leveraging this design, we retain the spatial indices corresponding to the bounding boxes selected after Non-Maximum Suppression (NMS). These indices allow us to directly retrieve and pool the associated feature vectors from the feature maps, eliminating the need for explicit RoI Align operations. Below, we explain how we further these features through Detection-Anchored Multi-scale Pooling (DAMP).

\paragraph{(1) Multi-scale gather with Gaussian neighbourhood.}
The anchor is snapped to each FPN level $i \in \{P3, P4, P5\}$ and a
Gaussian-weighted $3{\times}3$ neighbourhood is aggregated:
\begin{equation}
    \mathbf{g}_i = \sum_{\delta r,\, \delta c}
        \frac{e^{-(\delta r^2 + \delta c^2)}}{\mathcal{Z}}
        \cdot \mathrm{FPN}_i\!\left[r_i{+}\delta r,\; c_i{+}\delta c\right]
    \;\in \mathbb{R}^{C_i}
\end{equation}

\paragraph{(2) Projection and fusion.}
Each gathered vector is projected to dimension $D$, then the three levels are fused:
\begin{equation}
    \mathbf{o} = \mathbf{W}_{\text{fuse}}\,
    \mathrm{LN}\!\bigl(\mathbf{W}_{P3}\mathbf{g}_{P3}
                           \;\|\; \mathbf{W}_{P4}\mathbf{g}_{P4}
                           \;\|\; \mathbf{W}_{P5}\mathbf{g}_{P5}\bigr)
    \;\in \mathbb{R}^{D}
\end{equation}

This requires $N \times 9 \times 3$ gather operations vs.\ $N \times 7 \times 7 \times 3$ for RoI Align, a $5.4\times$ reduction with no
additional parameters in the gather stage. It is important to notice here that this operation is not necessarily more expressive than RoI Align because the Gaussian neighborhood is fixed across scale, which can introduce some noise.

\subsection{CARPE: Cross-Attention Rotary Prototype Embedding}

The core idea of the context modeling of REACT and PE-NET is prototype learning. Here we want to represent each relation triplet $(s,p,o)$ as a blend of visual and semantic signals, then score it against learnable predicate prototypes in a shared embedding space. In REACT, subject and object representations are fused linearly, then modulated by the spatial features. 
Instead of learning a fixed transformation from visual features to a relation-ready space, cross-attention lets each visual token query a bank of semantic prototypes (GloVe-lifted class embeddings) and selectively blend in whichever prototypes are most relevant. To implement this idea, we re-worked the entire prototype learning approach, starting with the $W_\text{sub}$, $W_\text{obj}$, $W_\text{pred}$ projections.

\paragraph{(1) Semantic lifting.}
REACT defines three independent two-layer ReLU MLPs $W_\text{sub}$, $W_\text{obj}$, $W_\text{pred}$:
\begin{equation}
    W_* : \mathbb{R}^{d_e} \to \mathbb{R}^{D/2} \to \mathbb{R}^D, \quad \sigma = \text{ReLU}.
\end{equation}
REACT\texttt{++} replaces all three with a single SwiGLU block \cite{shazeer2020glu} per modality:
\begin{equation}
    \mathrm{SwiGLU}(\mathbf{x}) = W_\text{out}\!\left(\mathrm{SiLU}(W_\text{gate}\,\mathbf{x}) \odot W_\text{val}\,\mathbf{x}\right),
\end{equation}
where $W_\text{gate}, W_\text{val} : \mathbb{R}^{d_e} \to \mathbb{R}^{d_h}$ and $W_\text{out} : \mathbb{R}^{d_h} \to \mathbb{R}^D$.
Sharing $W_\text{sub} = W_\text{obj}$ (tied weights) halves the lifting parameter count; the multiplicative gate eliminates the dead-neuron problem of ReLU.

\paragraph{(2) Visual–semantic fusion.}
REACT fuses visual and semantic signals via two explicit gate scalars:
\begin{align}
    g_s &= \sigma\!\left(W_{\text{gate\_sub}}\begin{bmatrix}\mathbf{t}_s\\\mathbf{v}_s\end{bmatrix}\right), \quad
    \mathbf{s} = W_\text{sub}\mathbf{t}_s + g_s \odot \mathbf{v}_s,
\end{align}
where $\mathbf{t}_s$ and $\mathbf{v}_s$ are the semantic and visual embeddings of the subject. An identical branch handles the object.

REACT\texttt{++} replaces this with a single \emph{cross-attention} module shared by both subject and object. Queries are visual features; keys and values are the full object prototype bank $\{\mathbf{c}^o_k\}$:
\begin{equation}
    \hat{\mathbf{s}} = \mathbf{x}_s + \mathrm{Attn}\!\left(\mathbf{x}_s,\,\{\mathbf{c}^o_k\},\,\mathbf{b}_s^\text{geo}\right),
\end{equation}
where $\mathbf{b}_s^\text{geo}$ is an additive geometry bias derived from the Rotary Position Embedding (see below). This lets the model select \emph{which} semantic prototypes to blend in via attention rather than a fixed gate, and is strictly more expressive at equal or lower parameter cost.

\paragraph{(3) Geometry encoding.}
REACT first uses two Conv blocks to learn the spatial representation of union boxes. These spatial features are included as additional information for the semantic prototype through a dedicated gate mechanism. Instead, REACT\texttt{++} introduces a learnable \emph{Geometry RoPE} \cite{su2024roformer} encoder applied in every forward pass:
\begin{equation}
    \mathbf{b}^\text{geo} = \mathrm{GeomRoPE}(\mathbf{b}) = \left[\sin(f_1(\mathbf{b})),\; \cos(f_2(\mathbf{b}))\right] \in \mathbb{R}^{d_\text{rope}},
\end{equation}
where $\mathbf{b} \in \mathbb{R}^9$ is the normalised box encoding (widths, heights, centres, area) and $f_1, f_2$ are small two-layer networks. $\mathbf{b}^\text{geo}$ is projected to one scalar per attention head and added to the query logits before softmax, enabling the attention pattern to vary smoothly with box position.

\paragraph{(4) Prototype bank.}
In REACT, predicate prototypes $\{\mathbf{c}_k\}$ are computed fresh each forward pass from the current embedding weights: $\mathbf{c}_k = \text{proj}(\text{ReLU}(W_\text{pred}\,\mathbf{t}_k))$. They carry no memory across iterations.

REACT\texttt{++} maintains an \emph{Exponential Moving Average} (EMA) shadow buffer alongside the gradient-trained prototype weights:
\begin{equation}
    \hat{\mathbf{c}}_k \leftarrow m\,\hat{\mathbf{c}}_k + (1-m)\,\bar{\mathbf{r}}_k^+, \quad m = 0.999,
\end{equation}
where $\bar{\mathbf{r}}_k^+$ is the mean normalised relation representation of class $k$ in the current batch. When the EMA buffer has been initialised for class $k$, the EMA prototype replaces the learnable weight; otherwise the learnable weight is used as fallback. This allows rare-class prototypes to remain stable across batches without gradient vanishing.



\subsection{Dynamic Candidate Selection}

Here, we introduce the DCS method to select an optimal number of proposal candidates at inference time. 
During training, the relation prediction stage takes as input a maximum amount of proposals $\theta$ ($\theta=100$). During evaluation, we uniformly sample the number of proposals $k$, starting from 0. Using the generated slope for each metric, the optimal DCS threshold is computed as : 
\(x_{\text{opt}} = \min \left\{ x \mid \left| f'(x) \right| < \epsilon \right\}.\) During inference, $x_{\text{opt}}$ is applied to control the maximum number of proposals to use as input for the relation prediction stage, successfully lowering the complexity of the task.

\section{Experiments}
\label{sec:Experiments-results}

To evaluate our approach, we propose a set of experiments on leading datasets. The goals of these experiments are threefold: (i) evaluate the Decoupled Two-Stage architecture against traditional Two-Stage frameworks; (ii) evaluate the new components of REACT++, and (iii) evaluate the Dynamic Candidate Selection (DCS) method for inference.

\begin{table}[t]
    \centering
    \caption[Results of experiments with YOLOV8 - full numbers]{Results on the PSG dataset. $\dagger$ represents evaluation with our DCS strategy. Bold and underlined represent the best and second best for each metric.}
    \setlength{\tabcolsep}{3pt}
    \renewcommand{\arraystretch}{1.6}
    \footnotesize
        \begin{tabular}{l|l|l|c|c|l|l|l|c}
            \hline
            \textbf{B} & \textbf{D} & \textbf{Relation Head} & \textbf{mR@20/50/100} & \textbf{R@20/50/100} & \textbf{F1@K} & \textbf{mAP} & \textbf{Lat.} & \textbf{Params}  \\ \hline
            \multirow{13}{*}{\rotatebox{90}{\textbf{CSPDarkNet-53}}} & \multirow{13}{*}{\rotatebox{90}{\textbf{YOLOV8m-DTS}}}  & 
            REACT++ (ours) $\dagger$ & \ \textbf{22.2} / \ \textbf{24.9} / \textbf{27.1} & \textbf{29.1} / \textbf{34.2} / \textbf{37.5} & \textbf{28.4} & \textbf{53.1} & \textbf{19.4} & \textbf{35.8M}\\
            & & REACT++ (ours) & \ \textbf{22.2} / \ \textbf{24.9} / \textbf{27.1} & \textbf{29.1} / \textbf{34.2} / \textbf{37.5} & \textbf{28.4} & 53.1 & 25.9 & \textbf{35.8M}\\
            & & REACT $\dagger$ \cite{Neau_2025_BMVC} & \ \underline{18.3} / \ 20.0 / 20.8 & 27.5 / 30.9 / 32.2 & \underline{23.9} & 53.1 & 23.0 & 43.3M\\
            & & REACT \cite{Neau_2025_BMVC} & 18.3 / \underline{20.1} / \underline{20.9} & 27.6 / 30.9 / 32.3 & 
            23.9 & 53.1 & 32.5 & 43.3M\\ 
            & & PE-NET $\dagger$ \cite{zheng2023prototype} & 17.1 / 19.0 / 19.8 & \underline{28.0} / \underline{31.3} / 32.9 & 23.2  & 53.1 & 51.7 & 187M \\ 
            & & PE-NET \cite{zheng2023prototype} & 17.1 / 19.0 / 19.9 & 28.0 / 31.3 / 33.0 & 23.2 & 53.1 & 103.0 & 187M \\
            & & GPS-NET $\dagger$ \cite{lin_gps-net_2020} & 10.9 / 12.6 / 13.5 & 26.3 / 30.0 / 31.8 & 17.3  & 53.1 & 27.8 & 45.3M \\
            & & GPS-NET \cite{lin_gps-net_2020} & 11.1 / 12.7 / 13.7 & 26.6 / 30.1 / 31.9 & 17.6 & 53.1 & 41.1 & 45.3M \\
            & & Motifs $\dagger$ \cite{zellers_neural_2018} & 10.1 / 11.6 / 12.5 & 25.7 / 28.7 / 30.3 & 16.2 & 53.1 & 22.4 & 52.9M \\
            & & Motifs \cite{zellers_neural_2018} & 10.1 / 11.6 / 12.5 & 25.7 / 28.7 / 30.4 & 16.2  & 53.1 & 35.8 & 52.9M \\
            & & VCTree $\dagger$ \cite{tang_learning_2019} & 10.7 / 12.2 / 12.9 & 22.1 / 25.3 / 26.8 & 16.1 & 53.1 & 183.2 & 268.0M \\ 
            & & VCTree \cite{tang_learning_2019} & 10.7 / 12.2 / 13.0 & 22.1 / 25.3 / 26.8 & 16.1 & 53.1 & 239.5 & 268.0M \\
            & & Transformer $\dagger$ \cite{tang_unbiased_2020} & 11.3 / 12.4 / 12.8 & 24.1 / 27.0 / 28.3 & 16.6 & 53.1 & \underline{21.1} & 97.9M \\ 
            & & Transformer \cite{tang_unbiased_2020} & 11.3 / 12.4 / 13.2 & 24.1 / 27.1 / 28.9 & 16.9 & 53.1 & 35.1 & 97.9M \\ \hline
            \multirow{5}{*}{\rotatebox{90}{\textbf{ResNeXt-101}}} & \multirow{5}{*}{\rotatebox{90}{\textbf{Faster-RCNN}}} & PE-NET \cite{zheng2023prototype} & 10.7 / 11.7 / 12.0 & 16.8 / 18.7 / 19.5 & 14.1  & 35.5 & 390.4 & 426.5M \\
            & & GPS-NET \cite{lin_gps-net_2020} & 8.2 / 8.8 / 9.2 & 15.7 / 18.1 / 19.2 & 11.7 & 31.6 & 313.9 & 391.6M \\ 
            & & Neural-Motifs \cite{zellers_neural_2018} & 8.9 / 9.6 / 9.8 & 16.9 / 18.7 / 19.5 & 12.5 & \underline{35.9} & 318.6 & 369.6M \\
            & & VCTree \cite{tang_learning_2019} & 7.9 / 9.1 / 9.6 & 18.0 / 20.7 / 22.3 & 12.4 & 32.6 & 519.5 & 365.9M \\
            & & Transformer \cite{tang_unbiased_2020} & 7.6 / 8.2 / 8.5 & 16.2 / 17.9 / 18.7 & 11.1 & 34.6 & 296.2 & 328.3M \\ \hline
            \multirow{2}{*}{\rotatebox{90}{\textbf{DETR}}} & \multicolumn{2}{c|}{EGTR \cite{im2024egtr}} & 12.0 / 14.5 / 16.6 & 24.9 / 30.3 / \underline{33.8} & 19.4 & 33.6 & 78.3 & \underline{42.5M} \\
            & \multicolumn{2}{c|}{RelTR \cite{cong_reltr_2022}} & 7.9 / 10.0 / 11.7 & 12.9 / 17.3 / 20.2 & 12.4 & 25.7 & 59.6 & 63.7M \\
            \hline
        \end{tabular}
    \label{tab:main_results_psg}
\end{table}

\subsection{Datasets}

In SGG, the most widely used dataset is Visual Genome \cite{krishna_visual_2017}. The commonly used split of the data is called VG150 and contains annotations from the top 150 object classes and top 50 predicate classes. However, VG150 suffers from severe biases \cite{yang_panoptic_2022,neau2023fine,hudsonGQANewDataset2019} and inherent issues such as the presence of ambiguous classes (e.g. the classes ``people", ``men", etc.) \cite{yang_panoptic_2022,liang_vrr-vg_2019}. Recently, two new datasets have been proposed with higher-quality annotations: the PSG \cite{yang_panoptic_2022} and IndoorVG \cite{neau2023defense} datasets. Both datasets have been carefully annotated to remove the presence of ambiguous classes or wrong annotations, which makes them a better choice for benchmarking our REACT++ architecture.

\subsection{Metrics.} 
Following previous work \cite{xu_scene_2017,tang_learning_2019}, we used the Recall@K (R@K) and meanRecall@K (mR@K) metrics to measure the performance of models. For both metrics, we evaluate the recall on the top $K$ ($k=[20,50,100]$) relations predicted, ranked by confidence.
The Recall@K evaluates the overall performance of a model on the selected dataset whereas meanRecall@K evaluates the performance on the average of all predicate classes, which is more significant for long-tail learning such as in the task of SGG. Both metrics are averaged in the F1@K metric as follows:
$F1@K = \frac{2 \times R@K \times mR@K}{R@K + mR@K}$
which efficiently represents the trade-off for the model performance between head and tail predictions. The Object Detection (OD) accuracy is measured using standard mAP@50 (mAP\textsuperscript{50}).

\subsection{Implementation details.} The latest YOLO versions (V8, V9, V10, V11, and V12) can be used interchangeably in our REACT++ architecture. For simplicity, we present experiments for the YOLOV8 medium version \cite{Jocher_Ultralytics_YOLO_2023} (YOLOV8m). YOLOV8m has been trained for object detection for 100 epochs on all datasets with a batch size of 16 and image size 640x640.
Relation prediction models have been trained for 20 epochs, batch size 8 with the SGD optimizer, and an initial learning rate of $1e^{-4}$.
For Dynamic Candidate Selection (DCS), we choose $\epsilon = 1e^{-5}$. Inference without DCS is performed with 100 proposals for all models.
In contrast to YOLO, models trained with Faster-RCNN use 600x1000 image sizes.
We fully re-trained seven different relation prediction models on the PSG and IndoorVG datasets: Neural-Motifs \cite{zellers_neural_2018}, VCTree \cite{tang_learning_2019}, Transformer \cite{tang_unbiased_2020}, GPS-Net \cite{lin_gps-net_2020}, PE-NET \cite{zheng2023prototype}, RelTR \cite{cong_reltr_2022}, and EGTR \cite{im2024egtr} following the provided codebases. All models are implemented without any re-weighting \cite{tengStructuredSparseRCNN2022b} or de-biasing strategy \cite{tang_unbiased_2020}. Latency is benchmarked with a batch size of 1\protect\footnotemark.
\footnotetext{Hardware used: \text{11th Gen Intel\textsuperscript{TM}} Core\textsuperscript{TM} i9-11950H @ 2.60GHz x 16, NVIDIA GeForce RTX 3080 Laptop GPU 16GB VRAM, 2 x 16GB 3200 MHz RAM.}

\subsection{Results}
In this section, we analyze the results obtained on the three different datasets. In the text, we refer to improvements as mean absolute percentage. 

\textbf{PSG dataset.} Results for the PSG dataset are displayed in \Cref{tab:main_results_psg}. We first compare the Decoupled Two-Stage (DTS) approach introduced in REACT with previous work.
By using DTS and YOLOV8, we improved the mAP by 54.37\% on the PSG dataset, compared to TS approaches with Faster-RCNN. Compared to OS approaches (EGTR \cite{im2024egtr} and RelTR \cite{cong_reltr_2022}), mAP is improved by 120\% on average, which demonstrates the inefficiency of such approaches for the OD task. We also experienced an average improvement of 58.47\% in F1@k by using our DTS method compared to the same approaches with Faster-RCNN. We observe that the gain in F1@k is not very strongly correlated to the gain in mAP between the Two-Stage and DTS approaches, as we measured a correlation coefficient of 0.75 between the two.
When we compare our REACT++ model with the original REACT, we observe an improvement of 5 points or 20\% in meanRecall@K on average. The same follows for the Recall@K, with 6 points improvements or +17.9\%. Those results are very strong and suggest the importance of the improvements made to the prototype learning module. Finally, we can observe that our REACT++ model is 20\% faster and consumes 17\% fewer parameters. 

\textbf{IndoorVG dataset.}
On the IndoorVG dataset, we observe an improvement of 43.4\% in mAP and 38.76\% in F1@K by comparing DTS and non-DTS variants of the same models. Our REACT++ model also outperforms REACT by a margin of 13\% in meanRecall@K (20.7 versus 18.0). However, REACT still outperforms REACT++ in Recall@K (30.9 versus 28.1). Since Recall is a head-tail biased metric, it does not hurt the expressiveness of the REACT++ model as shown by the F1@K (harmonic average of meanRecall@K and Recall@K).

\textbf{VG150 dataset.}
For a fair comparison, we also present the performance of the REACT++ model on the VG150 dataset, see \Cref{tab:results_vg150}. The performance of REACT in mAP is better by one point than the best approach EGTR. We believe that the low quality of the bounding box annotations and the ambiguous classes in VG150 are confusing the YOLOV8 detector, which leads to only a small improvement against Faster-RCNN.
We observe that REACT++ outperforms the original REACT and some Two-Stage approaches (Motifs, VCTree, and BGNN) but is not competitive with more recent approaches (PE-NET and SQUAT). We believe this is due to noisy feature representation. Because REACT++ has fewer parameters, it is more sensitive to noisy data.

\textbf{Latency.}
Regarding latency, we demonstrated an improvement of 84.99\% using DTS-YOLOV8m instead of Faster-RCNN, which is a considerable gap. Regarding the number of parameters, the DTS approach uses, on average 77.47\% fewer parameters than the baseline.
REACT++ outperforms REACT again by a strong margin, with 20\% less latency, dropping from 32.5 to 25.9.
By applying our DCS strategy, we gain an average of 66.5\% latency on all models (compared to the YOLOV8m-DTS versions) with a relative average loss of 1\% in F1@K. To our knowledge, REACT++ with DCS is the first model to attain a latency below 20ms for the task of Scene Graph Generation. 
REACT++ is the best overall in terms of parameter count (35.8M total).
By taking into account OD accuracy, RelPred accuracy, and latency, our REACT++ model is the best compromise. Even for the VG150 dataset, REACT++ is a better choice because no other two-stage approaches are close to real-time inference. On the other hand, when evaluated fairly, one-stage approaches are clearly behind in terms of mAP and F1@K on all datasets, and are not a good choice in comparison to REACT++.
\begin{table}[t]
\begin{minipage}{.49\textwidth}
\begin{center}
    \resizebox{\textwidth}{!}{%
    \renewcommand{\arraystretch}{1.2}
    \begin{tabular}{l|l|c|c|c|c}
        \hline
            \textbf{B} & \textbf{Relation Head} & \textbf{mR@K} & \textbf{R@K} & \textbf{F1@K} & \textbf{mAP\textsuperscript{50}} \\ \hline
            \multirow{7}{*}{\rotatebox{90}{\textbf{DTS}}} 
            & REACT++ & \textbf{20.7} & 28.1 & \textbf{23.9} & \textbf{37.2} \\
            & REACT & 18.0 & \textbf{30.9} & 22.8 & \textbf{37.2} \\
            & PE-NET & \underline{16.0} & 29.4 & \underline{20.7} & 37.2\\ 
            & GPS-NET & 9.9 & 29.2 & 14.8 & 37.2 \\ 
            & Neural-Motifs & 11.1 & \underline{30.5} & 16.2 & 37.2\\
            & VCTree & 14.9 & 18.9 & 16.7 & 37.2\\ 
            & Transformer & 12.3 & 28.4 & 17.2 & 37.2\\ \hline
            \multirow{5}{*}{\rotatebox{90}{\textbf{TS}}} & PE-NET & 13.8 & 27.1 & 18.1 & 25.2 \\ 
            & GPS-NET & 6.7 & 11.9 & 8.6 & 17.4 \\ 
            & Neural-Motifs & 9.8 & 15.2 & 11.9 & \underline{26.2}\\ 
            & VCTree & 11.0 & 14.7 & 12.6 & 25.5 \\ 
            & Transformer & 9.9 & 16.8 & 12.5 & 24.2 \\ \hline
            \multirow{2}{*}{\rotatebox{90}{\textbf{OS}}} & EGTR \cite{im2024egtr} & 7.1 & 10.6 & 8.5 & 14.7\\ 
            & RelTR \cite{cong_reltr_2022} & 7.7 & 11.4 & 9.2 & 14.1\\ \hline
        \end{tabular}
    }
    \end{center}
    \vspace{0.1cm}
    \footnotesize{(a) Results on IndoorVG. mR@K and R@K are average for $k=[20,50,100]$.}
    \label{tab:results_indoorvg}
\end{minipage}
\hfill
\begin{minipage}{.49\textwidth}
\begin{center}
    \resizebox{\columnwidth}{!}{%
        \renewcommand{\arraystretch}{1.2}
        \begin{tabular}{l|l|c|c|c|c}
        \hline
            \textbf{B} & \textbf{Model} & \textbf{mR@K} & \textbf{R@K} & \textbf{F1@K} & \textbf{mAP\textsuperscript{50}} \\ \hline
            \multirow{2}{*}{\textbf{DTS}}
            & REACT++ & 13.2 & 28.9 & 18.2 & \textbf{31.8} \\
            & REACT & 12.9 & 27.4 & 17.6 & 31.8 \\ \hline \multirow{4}{*}{\textbf{TS}} & PE-NET & 13.3 & 32.6 & \underline{18.9} & 29.2 \\ 
            & SQUAT \cite{jung2023devil} &  \textbf{15.3} & 26.7 & \textbf{19.4} & - \\
            & BGNN \cite{li_bipartite_2021} & 11.6 &  \underline{33.4} & 17.3 & 29.0 \\
            & VCTree & 7.2 & \textbf{33.9} & 11.8 & 28.1 \\ 
            \hline
            \multirow{4}{*}{\textbf{OS}} & EGTR \cite{im2024egtr} & 7.8 & 29.3 & 12.4 & \underline{30.8} \\ 
            & SGTR \cite{li_sgtr_2022} & \underline{13.6} & 26.5 & 18.0 & 25.4 \\ 
            & SS-RCNN \cite{tengStructuredSparseRCNN2022b} & 9.5 & 36.0 & 15.0 & 23.8 \\ 
            & RelTR \cite{cong_reltr_2022} & 8.0 & 24.3 & 12.0 & 26.4 \\ \hline
        \end{tabular}
    }
    \end{center}
    \vspace{0.1cm}
    \footnotesize{(b) REACT++ compared to previous work on VG150 \cite{xu_scene_2017,krishna_visual_2017} (reported results). mR@K and R@K are the average for $k=[50,100]$ as some approaches do not report $k=20$.}
    \label{tab:results_vg150}
\end{minipage}
\label{tab:results_vg150}
\caption{Our architecture REACT++ compared to previous work on: (a) results on IndoorVG  and (b) VG150 \cite{xu_scene_2017,krishna_visual_2017} (reported results).}
\end{table}

\section{Ablation Studies}
\label{sec:ablation}

This section presents our ablation studies in our REACT++ architecture regarding: (i) the introduced DAMP feature extractor; (ii) the impact of the global context with the AIFI block; (iii) the impact of the DCS strategy; and (iiii) further experiments with other YOLO variants.

\subsection{DAMP vs.\ RoI Align for Object Feature Extraction}

To isolate the effect of the DAMP box feature extractor, we trained 5 variants of the REACT\texttt{++} model for 10 epochs on the IndoorVG dataset. \textbf{RoiAlign} keeps the same algorithm used in REACT \cite{Neau_2025_BMVC}, borrowed from Motifs-TDE \cite{tang_unbiased_2020} and PE-NET \cite{zheng2023prototype}. In the REACT implementation, the RoI Align algorithm is used with a pooling size of $7 \times 7$.
\textbf{DA} (Detector-Anchored) replaces RoI Align with a single-vector gather using the YOLO detection-peak index $\iota_i$. This means we are directly extracting the exact feature vector that has been used to decode the corresponding bounding box and class label in the YOLO heads. This operation is almost free in terms of computing. \textbf{DAP} (Detector-Anchored with Pooling) applies the Detector-Anchored with a Gaussian-weighted neighbourhood on the same scale, with radius $r=1$ (9 features collected). \textbf{DAM} (Detector-Anchored with Multi-Scale) is aligning the index of the origin feature vector to the other level of the multi-scale feature map, without Gaussian-weighted neighbourhood, extracting 3 vectors per object. Finally, \textbf{DAMP} (Detector-Anchored with Multi-Scale Pooling) combines the two approaches to gather pooled features at different scales.

\begin{figure}
    \centering
    \includegraphics[width=\textwidth]{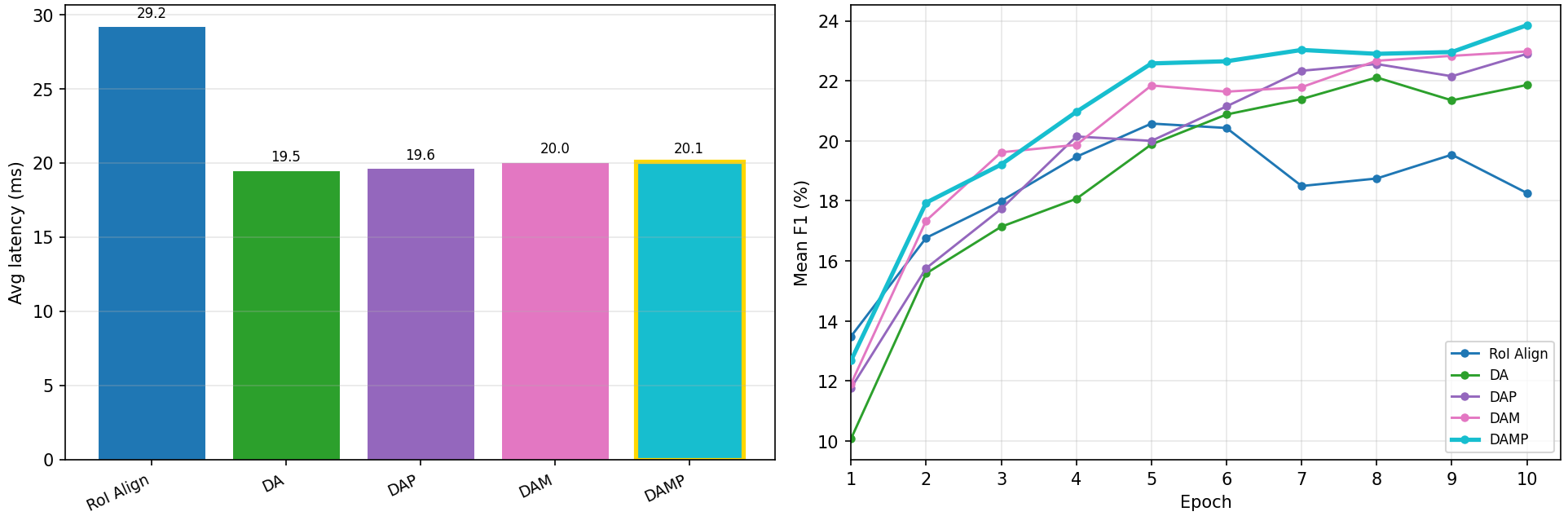}
    \caption{Left - latency comparison of the REACT++ model with the different Feature Extraction components. Right - evolution of the F1@K metric across different stages of training, for the same components.}
    \label{fig:feature_extract_ablation}
\end{figure}

\Cref{fig:feature_extract_ablation} summarizes accuracy on the test split and latency profiled over 200 validation images with 20 warmup steps. For simplicity, the latency only represents the \textit{forward pass} of the model. 
First, we can notice in \Cref{fig:feature_extract_ablation} (left) the consequent latency gap between the RoI Align approach and direct feature extraction by indexing using DA. The gap is 9.3ms on average, or 32\%.
DA reduces ROI Heads latency by 0.6\,ms (3\%) compared to DAMP but with the cost of 1.9 points or 8\% in F1@K (23.8 versus 21.9). Pooling features with Gaussian neighbors (DAP) seems to help with an improvement of 1 point in F1@K compared to DA, with almost no cost in latency. Finally, DAM also introduces almost no additional cost in latency but slightly less gain (+0.5 points against DA). The full DAMP is preferred in terms of accuracy (+0.9 point F1@K), with a relatively good latency compared to the original RoI Align. In \Cref{fig:feature_extract_ablation} (right), we observe that the evolution of DAMP in F1@K stays consistent across epochs, enforcing the stability of the approach.

\subsection{Impact of the AIFI block for global context}

In this ablation study, we analyzed the global context branch of REACT++ on the IndoorVG dataset (best checkpoint over 10 epochs). We used two variants: the baseline REACT++ model with AIFI for global context and a REACT model with no global context branch. Both models use DAMP box features and the CARPE predictor with geometryRoPE bias. Results are in \Cref{tab:context_ablation}. We can see that the AIFI global context branch is significantly contributing to the overall accuracy, with an improvement of +0.42 points in F1@K (1.8\%). Even if the contribution is marginal, it can be important on the tail classes, as shown by the mR@K. Regarding overall parameters count and latency, the model without AIFI outperforms.
  
\begin{table}[h]
\centering
    \caption{%
      Ablation of the AIFI component of REACT++. \emph{Params} counts only unfrozen relation-head parameters. Metrics are averaged over @20, @50, and @100 thresholds. Latency is the median \emph{model forward} time over 200 images.
    }
\setlength{\tabcolsep}{8pt}
\begin{tabular}{l ccc cc}
\toprule
Method
  & R@$K$ 
  & mR@$K$ 
  & F1@$K$ 
  & Params (M)
  & Latency (ms) \\
\midrule
  Global context (AIFI)
  & \textbf{28.13} & \textbf{20.70} & \textbf{23.85} & 9.82 & 20.76 \\
  No global context
  & 27.84 & 20.23 & 23.43 & \textbf{7.85} & \textbf{17.94} \\
\midrule
$\Delta$ (AIFI $-$ no context)
  & $+$0.29 & $+$0.47 & $+$0.42 & $-$1.97 & $-$2.82 \\
\bottomrule
\end{tabular}
\medskip
\label{tab:context_ablation}
\end{table}

\subsection{Latency and Number of Proposals: the DCS case} 

In this section, we measured the importance of the Dynamic Candidate Selection method for low latency during inference. To do so, we ran experiments with different numbers of proposals.
In fact, the second stage of the REACT++ architecture takes as input a fixed number of $n$ proposals as node candidates. There could be a maximum of $n \cdot (n-1)$ possible pairs in the graph; thus, when doing matching, the computational complexity is supposed to scale accordingly. To demonstrate this hypothesis, we evaluated the performance of the REACT++ model in both latency and accuracy for different numbers of input proposals $k=\{10,150\}$ with DTS-YOLOV8m. For all experiments, we ranked proposals by confidence of the backbone and selected the top $k$.
\begin{figure}[t]
    \centering
    \begin{subfigure}{.49\textwidth}
    \centering
        \includegraphics[width=\textwidth]{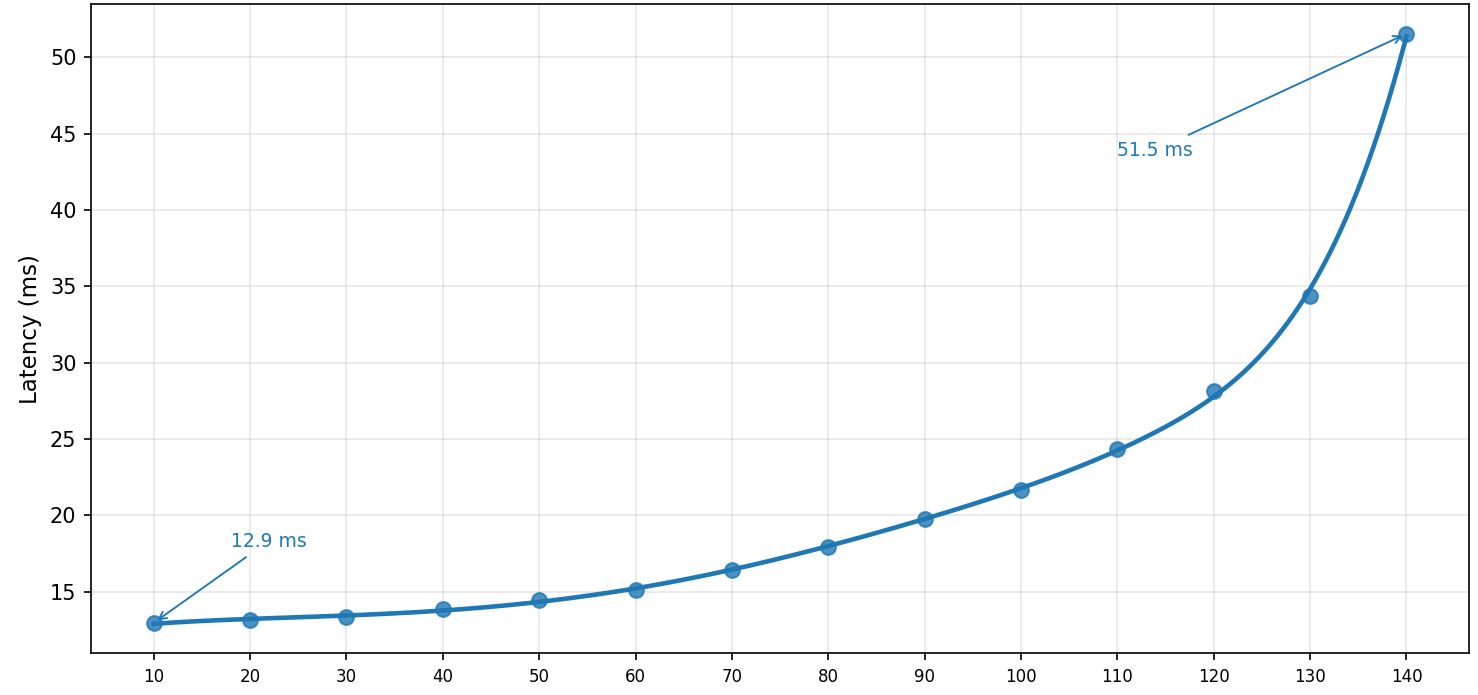}
    \end{subfigure}
    \begin{subfigure}{.49\textwidth}
    \centering
        \includegraphics[width=\textwidth]{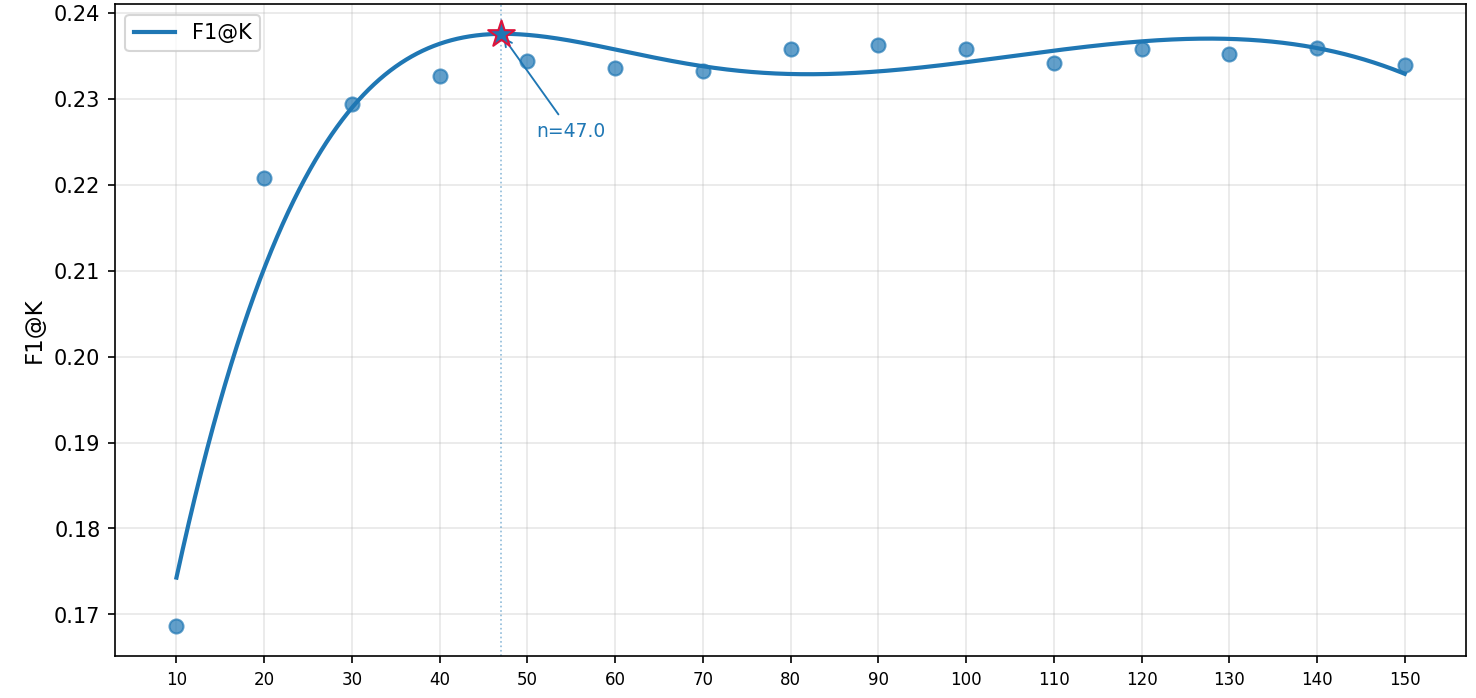}
    \end{subfigure}
    \caption{Left: Latency for the REACT++ model using a different number of proposals per image, with batch size 1. Right: Average F1@k for the REACT++ model with different number fo proposals.}
    \label{fig:num_proposals}
\end{figure}
The results of these experiments are shown in \Cref{fig:num_proposals}, where we observe a constant improvement, with an important increase after 100 proposals. Even with 140 proposals, the latency of the REACT++ model stays under all previous Two-Stage and One-Stage approaches. With 130 proposals, the latency is similar to that of the original REACT with 100 proposals (32.5ms).
The evaluation of the REACT++ model under different sampling conditions with our DCS strategy is represented in \Cref{fig:num_proposals} which shows that the optimal value was found at 23.89 for $k=47$, just before the performance started to be saturated.

\begin{figure}
    \centering
    \includegraphics[width=\textwidth]{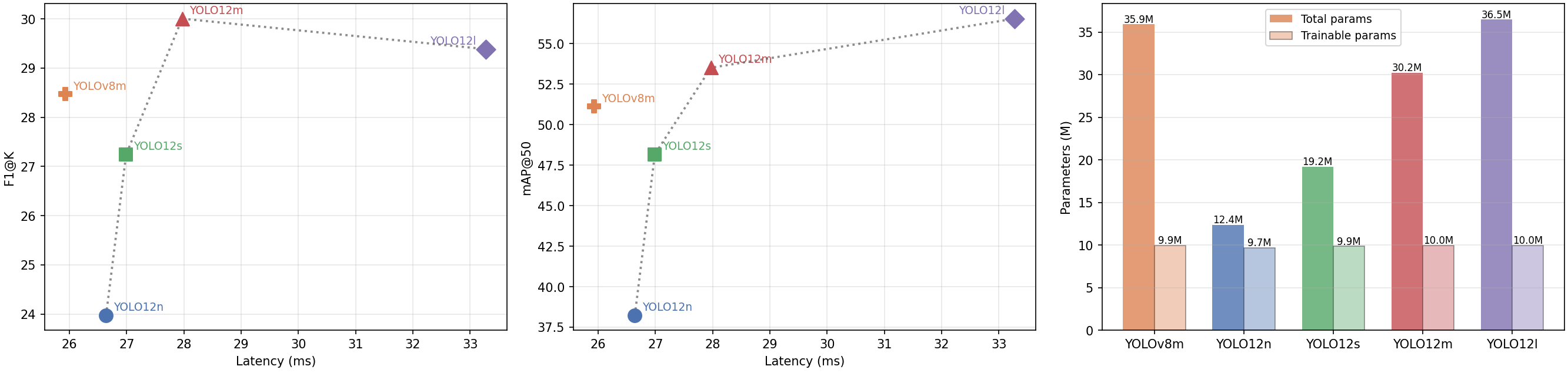}
    \caption{Average Recall@K, meanRecall@K, and mAP@50 performance for REACT++ against the corresponding latency using different variants of the YOLO12 model \cite{tian2025yolov12}.}
    \label{fig:pe_net_variants}
\end{figure}

\subsection{Scaling REACT++} 

To better evaluate the impact of the object detector accuracy on the relation prediction stage, we ran a set of experiments using different variants of YOLO with the REACT++ model for Scene Graph Generation. To demonstrate that REACT++ can be applied with any YOLO version (V8 / 9 / 10 / 11 / 12) and any variants (nano, small, medium, large), we decided to run experiments on the YOLO12 family \cite{tian2025yolov12}, in addition to the YOLOV8-m model used previously.
Here, we trained the REACT++ model with the same hyperparameters as before, only switching the object detector backbone and modifying the input size of the visual features used for relation prediction to match the different output channel sizes of each YOLO12 variant. The results are displayed in \Cref{fig:pe_net_variants}. The performance of the larger model, YOLO12-L, is slightly worse in F1@K than the medium YOLO12-M. A situation already observed in the original REACT work with the YOLOV8 family \cite{Neau_2025_BMVC}. We believe that the higher number of dimensions in the features extracted from the backbone could lead to more difficulties for the relation prediction training to converge.
Interestingly, we observed that REACT++ with YOLOV8-medium is faster than YOLO12-nano and YOLO12-small, which are smaller in terms of parameters (see \Cref{fig:pe_net_variants} - right).
In addition, we observed a consistent increase in latency with the size of the model in the YOLO12 family. 
In terms of the overall performance, the YOLO12-medium version seems to be the best, surpassing largely the results obtained with YOLOV8m in \Cref{tab:main_results_psg}. We believe that the Area Attention mechanism of YOLO12 leads to better feature representations, which helps the REACT++ model to disambiguate between fine-grained predicates. REACT++ with YOLO12m is the first model to ever attain 30 in F1@K on the PSG dataset.







\section{Conclusion}
\label{sec:conclusion}

This paper proposed REACT++, a novel implementation of a real-time Scene Graph Generation model that attains competitive performance with a latency of 25.9ms. Building on REACT, we proposed a new architecture with Decoupled Two-Stage SGG (DTS), aiming at improving real-time object detectors for SGG. Using this architecture, we replaced the Faster-RCNN backbone with YOLO for object detection and a newly introduced Detection-Anchored Multi-scale Pooling (DAMP) for feature extraction. We also proposed a new module for global context using a low-cost AIFI block to infuse scene dynamics to the prototype embedding representation, improving relation prediction. In addition, we proposed a new Cross-Attention Rotary Prototype Embedding (CARPE) module, which improves prototype embedding with dedicated subject/object cross-attention.
We also proposed a new inference method, Dynamic Candidate Selection (DCS), to further reduce latency by 66.5\% on average without compromising performance. 
Thus, our architecture REACT++ combines the DTS architecture with efficient prototype learning in order to decrease latency and rise accuracy.

Our main finding is that the performance and latency of SGG models are heavily correlated with the quality and quantity of object proposals generated by the object detector. However, regarding both overall performance and latency, a trade-off can be found by using our REACT++ model.
Another finding is the bottleneck of the ROI Align module, which could be easily replaced with a direct index-based pooling on one-stage object detectors such as YOLO.
Future work will consider implementing the REACT++ model in constraint settings such as embodied agent navigation or reasoning. Due to its small size, REACT++ can be embedded onboard robotic platforms and provide reliable and fast predictions that can foster reasoning.

\section*{Acknowledgements}
\noindent This work has been supported by the Knut and Alice Wallenberg foundation and the Wallenberg AI, Autonomous Systems and Software Program (WASP). The computations were enabled by resources provided by the National Academic Infrastructure for Supercomputing in Sweden (NAISS), partially funded by the Swedish Research Council through grant agreement no. 2022-06725.

\section*{Declarations}

\textbf{Data availability:} All code and data used in this paper will be made available on request.

\clearpage

\begin{appendices}
\section{Hyperparameters}\label{secA1}

\begin{table}[h!]
    \centering
    \caption{Main hyperparameter differences between REACT and REACT++ on PSG.}
    \setlength{\tabcolsep}{5pt}
    \renewcommand{\arraystretch}{1.4}
    \footnotesize
    \begin{tabular}{l|c|c}
        \hline
        \textbf{Hyperparameter} & \textbf{REACT} & \textbf{REACT++} \\ \hline
        MLP hidden dim          & 1024                            & 512 \\
        Loss function           & CrossEntropy                    & Logit Adjustment + Focal Loss \\
        Optimizer               & SGD  & AdamW  \\
        Learning Rate & $10^{-2}$ & $10^{-4}$ \\
        Scheduler             & ReduceLROnPlateau               & CosineAnnealing \\
        Gradient accumulation   & 1                               & 4 \\
        \hline
    \end{tabular}
    \label{tab:react_vs_reactpp_hparams}
\end{table}

\section{Ablation: DAMP}

\begin{table}[h!]
    \centering
    \caption{Ablation study on the box feature extractor on the IndoorVG dataset.
    DA: Detection-Anchored.
    DAM: DA + multi-scale (P3+P4+P5, no gather).
    DAMP: DAM + $3{\times}3$ Gaussian-weighted neighbourhood gather (ours).}
    \label{tab:ablation_extractor}
    \setlength{\tabcolsep}{6pt}
    \renewcommand{\arraystretch}{1.4}
    \footnotesize
    \begin{tabular}{l|c|c|c|c}
        \hline
        \textbf{Extractor} & \textbf{mR@K} & \textbf{R@K} & \textbf{F1@K} & \textbf{Lat. (ms)} \\ \hline
        RoIAlign (baseline) &  4.9 & 21.1 &  8.0 & 29.2 \\
        DA                  & 18.2 & 27.4 & 21.9 & \textbf{19.5} \\
        DAP                 & 19.4 & 27.9 & 22.9 & 19.6 \\
        DAM                 & 19.1 & 27.0 & 22.4 & 19.6 \\
        \textbf{DAMP (ours)}& \textbf{20.7} & \textbf{28.1} & \textbf{23.8} & 20.8 \\
        \hline
    \end{tabular}
    \vspace{0.3em}
\end{table}

\clearpage

\section{Ablation: Scaling REACT++}

\begin{table}[h]
    \centering
    \caption{Comparison of REACT++ with different YOLO backbone sizes on the PSG dataset.}
    \setlength{\tabcolsep}{4pt}
    \renewcommand{\arraystretch}{1.5}
    \footnotesize
    \begin{tabular}{l|c|c|c|c|c|c}
        \hline
        \textbf{Model} & \textbf{mR@20/50/100} & \textbf{R@20/50/100} & \textbf{F1@K} & \textbf{mAP\textsuperscript{50}} & \textbf{Lat. (ms)} & \textbf{Params} \\ \hline
        YOLOv8m  & 22.2 / 24.9 / 27.0 & 29.1 / 34.2 / 37.5 & \underline{28.5} & 51.1 & \textbf{25.9} & 35.9M \\
        YOLO12n  & 17.1 / 20.9 / 22.4 & 25.6 / 30.0 / 33.2 & 24.0 & 38.2 & \underline{26.6} & \textbf{12.4M} \\
        YOLO12s  & 21.2 / 23.9 / 25.8 & 27.7 / 32.7 / 36.0 & 27.2 & 48.2 & 27.1 & \underline{19.2M} \\
        YOLO12m  & \textbf{23.5} / \textbf{26.4} / \textbf{28.3} & \textbf{30.9} / \textbf{36.1} / \textbf{39.3} & \textbf{30.0} & \underline{53.5} & 28.0 & 30.3M \\
        YOLO12l  & \underline{23.6} / \underline{26.1} / \underline{27.9} & \underline{29.5} / \underline{34.9} / \underline{37.8} & 29.4 & \textbf{56.6} & 33.3 & 36.5M \\
        \hline
    \end{tabular}
    \label{tab:psg_yolo_variants}
\end{table}

\end{appendices}


\clearpage

\bibliography{SGG_bilblio}

\end{document}